\documentclass[10pt,twocolumn,letterpaper]{article}
\usepackage[accsupp]{axessibility}
\usepackage{iccv}
\usepackage{cite}
\usepackage{times}
\usepackage{epsfig}
\usepackage{graphicx}
\usepackage{amsmath}
\usepackage{amssymb}
\usepackage{threeparttable}
\usepackage{multirow}
\usepackage{booktabs}
\usepackage{paralist}
\usepackage{bbding}
\usepackage{color}
\usepackage{enumerate}
\usepackage{algorithm}
\usepackage{algorithmic}
\usepackage{colortbl}
\usepackage{csquotes}
\usepackage[normalem]{ulem}
\usepackage{fontawesome}

\usepackage[pagebackref=true,breaklinks=true,colorlinks,bookmarks=false]{hyperref}

\usepackage[capitalize]{cleveref}
\crefname{section}{Sec.}{Secs.}
\Crefname{section}{Section}{Sections}
\Crefname{table}{Table}{Tables}
\crefname{table}{Tab.}{Tabs.}

\definecolor{mygray}{gray}{.9}
\definecolor{myred}{rgb}{0.8,0.2,0}
\definecolor{myblue}{rgb}{0,0,1.0}

\iccvfinalcopy 

\def\iccvPaperID{7222} 

\ificcvfinal\fi

\begin{document}

\title{Multimodal Variational Auto-encoder based Audio-Visual Segmentation}

\author{Yuxin Mao\textsuperscript{1} \space\space\space\space Jing Zhang\textsuperscript{2} \space\space\space\space Mochu Xiang\textsuperscript{1} \space\space\space\space Yiran Zhong\textsuperscript{3} \space\space\space\space Yuchao Dai\textsuperscript{1}$^{\dagger}$ \\
\textsuperscript{1}\normalsize Northwestern Polytechnical University \& Shaanxi Key Laboratory of Information Acquisition and Processing \\ \space\space\space\space \textsuperscript{2}\normalsize Australian National University
\space\space\space\space \textsuperscript{3}\normalsize Shanghai AI Laboratory \\
\space\space\space\space \normalsize\textcolor{black}{\faGithub}~\href{https://github.com/OpenNLPLab/MMVAE-AVS}{\textcolor{blue}{https://github.com/OpenNLPLab/MMVAE-AVS}} 
\space\space\space\space\space\space\space\space \normalsize\textcolor{black}{\faGlobe}~\href{https://npucvr.github.io/MMVAE-AVS}{\textcolor{blue}{https://npucvr.github.io/MMVAE-AVS}}
}

\maketitle
\ificcvfinal\thispagestyle{empty}\fi

\noindent\let\thefootnote\relax\footnotetext{{${\dagger}$ Corresponding author \tt(daiyuchao@gmail.com).}}
\noindent\let\thefootnote\relax\footnotetext{This work was done when Yuxin Mao was an intern at Shanghai AI Laboratory.}
\begin{abstract}
We propose an Explicit Conditional Multimodal Variational Auto-Encoder (ECMVAE) for audio-visual segmentation (AVS), aiming to segment sound sources in the video sequence.
Existing AVS methods focus on implicit feature fusion strategies, where models are trained to fit the discrete samples in the dataset. With a limited and less diverse dataset, the resulting performance is usually unsatisfactory.
In contrast, we address this problem from an effective representation learning perspective, aiming to model the contribution of each modality explicitly.
Specifically, we find that audio contains critical category information of the sound producers, and visual data provides candidate sound producer(s). Their shared information corresponds to the target sound producer(s) shown in the visual data.
In this case, cross-modal shared representation learning is especially important for AVS.
To achieve this, our ECMVAE factorizes the representations of each modality with a modality-shared representation and a modality-specific representation.
An orthogonality constraint is applied between the shared and specific representations to maintain the \emph{exclusive} attribute of the factorized latent code. 
Further, a mutual information maximization regularizer is introduced to achieve extensive exploration of each modality.
Quantitative and qualitative evaluations on the AVSBench demonstrate the effectiveness of our approach, leading to a new state-of-the-art for AVS, with a 3.84 mIOU performance leap on the challenging MS3 subset for multiple sound source segmentation. 
\end{abstract}

\section{Introduction}
\label{sec:intro}

Audio-visual data can work collaboratively towards a better perception of the scene. 
The audio-visual segmentation (AVS)~\cite{zhou_AVSBench_ECCV_2022, mao2023contrastive} task aims to segment the objects from the video sequence that producing the sound in the audio. 
On the one hand, the audio data provides category information for the localization of the object in the video. On the other hand,
the visual data provides a sound producer pool with precise structure information of the foreground (sound producer(s)). 
Different from conventional multimodal settings, where each modality can be used individually for the target task, audio in AVS task serves as \enquote{command} to localize and segment the sound producer(s) from the visual data. In this case, the contribution of audio should be extensively explored for accurate foreground segmentation.


The baseline model~\cite{zhou_AVSBench_ECCV_2022} focuses on implicit feature fusion via audio-visual cross attention. It purely relies on fitting the discrete samples in the dataset. 
Although reasonable performance is obtained, there are no constraints to guarantee the contribution of each modality, making it
hard to decide if the audio data is effectively used,
as the model can directly regress the final segmentation maps by only taking the video as input. 
Fortunately, we discover that each modality for AVS contains both shared and specific information.
For example, the audio data includes both the information from the sound producers and the background noise, while the visual data shows the appearance of the entire scene, where
the sound producers only take a small portion of it.
In our specific task setting, we find modality factorization is suitable to model both the modality-shared representation, \ie~information of the sound producers, and the modality-specific representation toward a better understanding of the contribution of each modality.

The straightforward solution to learn the feature representation
of the input data 
is through an auto-encoder (AE) framework. However, AE is mainly used for data compression, as the learned feature space is not continuous, which cannot provide a rich semantic correlation of the data. Differently, with latent space regularization, \eg~the latent space is assumed to be Gaussian in variational auto-encoder (VAE)~\cite{VAE1}, VAE obtains semantic meaningful latent space, which is continuous, and it is also the basic requirement for reliable latent space factorization.

To learn the semantic correlated feature representation of the AVS data, we propose an Explicit Conditional Multimodal Variational Auto-Encoder (ECMVAE) for audio-visual segmentation to learn both the \textit{shared} and the \textit{specific} representation in the latent space of each modality. Our model is built upon a multimodal variational auto-encoder~\cite{multimodal_vae,Multimodal_Generative_Learning_Jensen_Shannon_Divergence}, with the Jensen-Shannon divergence to achieve a trade-off between sampling efficiency and sample quality. Based on the latent space factorization, we impose constraints for the shared and specific representations to explicitly maximize the contribution of each modality.

Specifically, we first assume that one latent code of the factorized representation should contain independent information compared to others. Furthermore, for the fused representation, we further claim that it should be more informative for the target task compared with each modality.
To achieve the former, we propose an information orthogonality constraint between the factorized representations of each modality to ensure that the modality-shared and modality-specific representations capture different aspects of the audio-visual input.
For the latter, we fuse the factorized representations of each modality to construct a fused space. Then we introduce a mutual information maximization regularizer between the fused representations of each modality to extensively explore the contribution of each modality.
Extensive experimental results demonstrate that our ECMVAE achieves \emph{state-of-the-art} AVS performance.
Our pipeline achieves a 3.84 mIOU improvement for the challenging multiple sound source segmentation. 



We summarize our main contributions as:
\begin{compactitem}
    \item An explicit semantic correlated feature representation
    learning framework for audio-visual segmentation is proposed with latent space factorization to capture both the modality-shared and specific representations.
    \item Based on the latent space factorization, we introduce a unimodal orthogonality constraint between the shared and specific representations and the cross-modal mutual-information maximization regularizer to extensively explore the contribution of each modality.
    \item State-of-the-art segmentation performance is achieved, showing both the effectiveness of each module and the contribution of each modality.
\end{compactitem}

\section{Related Work}
\noindent\textbf{Audio-Visual Segmentation.}
The Audio-Visual Segmentation (AVS) task is newly proposed,
aiming to localize the sound producers with
pixel-wise segmentation masks.
Zhou \etal~\cite{zhou_AVSBench_ECCV_2022} propose an AVSBench dataset for audio-visual segmentation and provide a simple baseline based on temporal pixel-wise audio-visual interaction (TPAVI), which is a cross-modal attention~\cite{vaswani_attention_is_2017_NIPS} based fusion strategy.
The other audio-visual collaboration tasks can be classified as audio-visual correspondence (AVC)~\cite{arandjelovic_look_listen_and_learn_CVPR_2017, arandjelovic_objects_that_sound_ECCV_2018}, event localization (AVEL)~\cite{lin_avel1cite_icassp_2019, lin_avel2cite_icassp_20120, tian_avel3cite_eccv_2018, zhou_avel4cite_cvpr_2021, xuan_avel_cross_atten1_aaai_2020, yu_avel_cross_atten1_mm2022}, event parsing (AVP)~\cite{tian_avpcite1_eccv_2020, wu2021_avpcite2_cvpr_2021, lin_avpcite3_2021exploring},~\etc These methods require the fusion of audio and visual signals. Such as audio-visual similarity modeling by computing the correlation matrix~\cite{arandjelovic_look_listen_and_learn_CVPR_2017, arandjelovic_objects_that_sound_ECCV_2018}, audio-visual cross attention~\cite{xuan_avel_cross_atten1_aaai_2020, yu_avel_cross_atten1_mm2022, Liang_2021_ICCV_Asynchronous_Multimodal}, audio-guided Grad-CAM~\cite{qian_audio_guided_cam_eccv_2020}, or using a multimodal transformer for modeling the long-range dependencies between elements across modalities directly~\cite{tsai2019MULT, attention_bottleneck_multimodal_fusion}.
However, the challenge and uniqueness of the AVS task are how to map the audio signals to \emph{fine-grained} visual cues, \ie per-pixel segmentation maps. This will rely on reliable modeling of visual and audio signals, as well as more effective fusion strategies.

\noindent\textbf{Multimodal Variational  Auto-encoders.}
The Multimodal Variational Auto-encoders (MVAEs)~\cite{multimodal_vae, joy2022learning_multimodal_vae, suzuki_jmvae_2016, mathieu2019disentangling, ma_smil_aaai_2021} are a type of latent variable generative model
to learn more
generalizable representations from diverse modalities. To achieve this, the core of MVAE is the joint distribution estimation. The conventional unimodal VAEs~\cite{VAE1,cvae_nips} are optimized by maximizing the evidence lower bound (ELBO), which includes a reconstruction term and the Kullback-Leibler (KL) divergence term to measure the divergence from the variational posterior to the prior distribution of the latent variable. In a multimodal setting, the KL divergence is defined between the joint posterior and joint prior across the modalities, which is often estimated by the product of experts (PoE)~\cite{product_of_expert,Generalized_Product_of_Experts} or the mixture of experts (MoE)~\cite{shi2019variational}. 
Based on such a prerequisite, many works extend the basic MVAE definition, such as missing modality handling~\cite{ma_smil_aaai_2021}, latent space modeling~\cite{shi2019variational, mathieu2019disentangling}, effective divergence modeling~\cite{Multimodal_Generative_Learning_Jensen_Shannon_Divergence}, \etc However, previous MVAE based frameworks essentially focus on the multimodal image generation task.
In this work, we bring the MVAE to the AVS task and propose the conditional version~\cite{cvae_nips, zhang_ucnet_CVPR_2020} of MVAE with practical multimodal information constraints for segmentation.

\noindent\textbf{Mutual Information Estimation.}
Mutual Information (MI) captures the nonlinear statistical dependencies between variables, acting as a measure of actual dependence~\cite{Equitability_mutual_information}. Specifically, for a pair of random variables $X$ and $Y$, their MI $I(X;Y)$ is defined as the KL divergence of the joint distribution $p_{(X,Y)}$ from the product of the marginal distributions $p(X)$ and $p(Y)$, which measures the shared information between $X$ and $Y$.
Although mutual information is simple in formation, as the log density ratio between the joint distribution $p_{(X,Y)}$ and product of marginals $p(X)\otimes p(Y)$ is intractable, it is usually estimated~\cite{Estimation_Conditional_Mutual_Information_using_MinMax_formulation,ccmi_classifier_based_mutual_information,Conditional_Mutual_Information_Estimation_Feature_Selection} instead of computed directly, leading to both MI maximization with a lower bound~\cite{poole2019variational,ba_mmm_nips03} and MI minimization with an upper bound~\cite{cheng2020club}. 
The MI maximization is usually applied for effective self-supervised representation learning~\cite{sanchez2020learning,xin2021disentangled,gao2021information,soleymani2021mutual,Disentangled_Representation_Learning_with_Wasserstein_Total_Correlation} for the unimodal data to guarantee reliable feature representation.
For the multimodal tasks, when each modality of data contains partial information of the target, both MI maximization and minimization can be applied~\cite{Mao_Yan_Guo_Ye_2021,liao2021multimodal,MIMF_Mutual_Information_Driven_Multimodal_Fusion,Improving_Multimodal_fusion_via_Mutual_Dependency_Maximisation,Modality_Specific_Representations_MSA,factorzed_multimodal_representation,multimodal_fusion_msa,Private_Shared_Disentangled_Multimodal_VAE,Self_Supervised_Disentanglement_Modality_Specific_Shared_Factors_Improves_Multimodal_Generative_Models}, where the former aims to explore the task-driven feature across the modalities, and the latter is designed to explore the complementary information of different modalities.
Extensive researches show that effective MI optimization can not only lead to informative representation learning~\cite{mutual_information_estimation_representation_learning,Improved_Mutual_Information_Estimation,mine_mutual_information,Wasserstein_Dependency_Measure_for_Representation_Learning,Kemertas_2020_CVPR_RankMI,do2021clustering,bachman2019learning,hjelm2018learning} but also is beneficial for achieving adversarial robustness~\cite{Adversarial_Robustness_via_Mutual_Information_Estimation}.

\begin{figure*}[!htp]
\begin{center}
\includegraphics[width=0.95\linewidth]{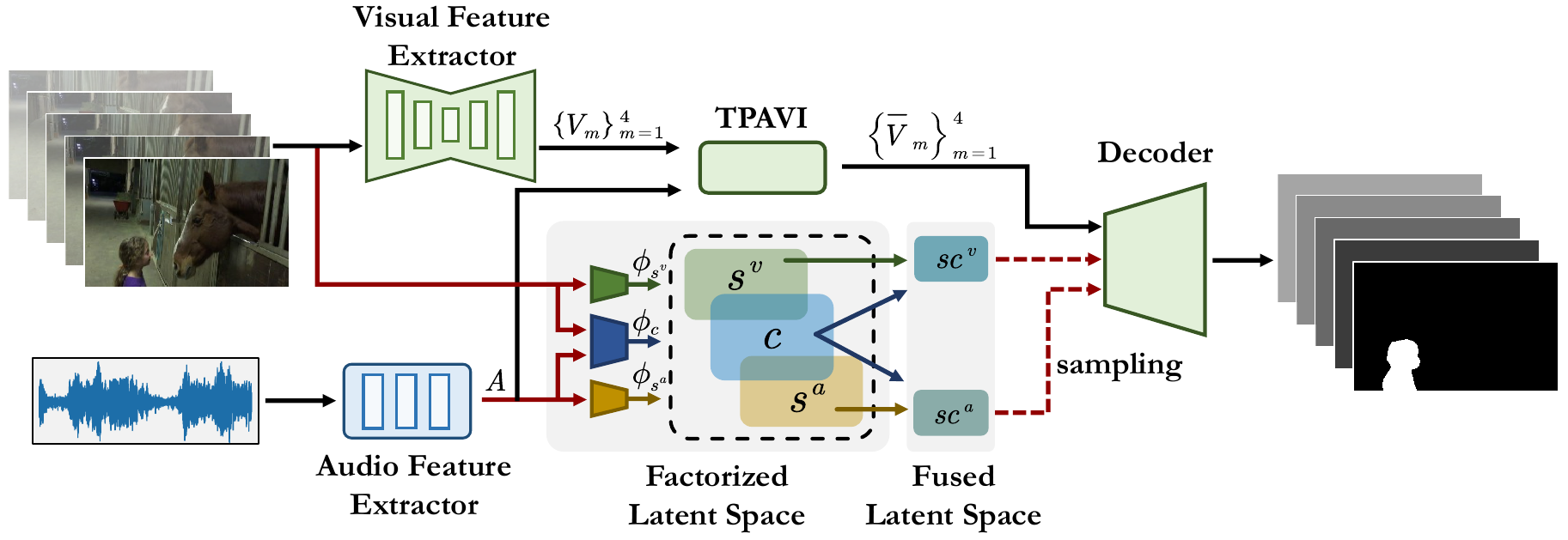}
\end{center}
\vspace{-0.0mm}
\caption{\textbf{Overview of the proposed ECMVAE for audio-visual segmentation}. The feature extractors are used to extract backbone features for the two modalities. 
We also design three latent encoders $\phi_{s^v}, \phi_{s^a}, \phi_{c}$ to achieve latent space factorization and obtain both task-driven shared representation $(c)$ and modality-related specific representation $(s^a, s^v)$, achieving explicit multimodal representation learning. The decoder is introduced to obtain the final segmentation maps, indicating the sound producers of the audio-visual data.}
\vspace{-2.0mm}
\label{fig:model_overview}
\end{figure*}

\section{Explicit Conditional Multimodal Learning}
We denote the input data of our used AVSbench dataset~\cite{zhou_AVSBench_ECCV_2022} is $X\!=\!\{\{x^v_t\}_{t=1}^T,x^a\}$, \ien~the visual $\{x^v_t\}_{t=1}^T$ for $T$ non-overlapping yet continuous frames, audio $x^a$ of the current clip. $y=\{y_t\}_{t=1}^T$ is the output, \ien~the segmentation maps (we omit $t$ for clear presentation). The goal of audio-visual segmentation is to segment the objects from the video $\{x^v_t\}_{t=1}^T$ that produce the sound
shown in
the audio 
$x^a$.
We introduce an Explicit Conditional Multimodal Variational Auto-Encoder (ECMVAE) using Jensen-Shannon divergence (Sec.~\ref{sec:cvae_preliminary})
via latent space factorization (Sec.~\ref{sec:latent_space_factorization}) to effectively model the shared representation between the two modalities (Sec.~\ref{orthogonality_constraint} and Sec.~\ref{shared_rep_regularization}) for audio-visual segmentation. The overview of the proposed ECMVAE is shown in Fig.~\ref{fig:model_overview}.

\subsection{Prerequisite}
\label{sec:cvae_preliminary}
\noindent\textbf{Conditional Variational Auto-encoder.} We begin our prerequisite with the definition of the conditional variational auto-encoder (CVAE)~\cite{VAE1,cvae_nips}, which contains a generative process and an inference process.
The generative process is to draw the latent variable $z$ from the prior distribution $p_\theta(z|X)$ with a given $X$, and generate the output via $p_\theta(y|X,z)$, where in our case $X$ and $y$ in the following derivations are $X=\{\{x^v_t\}_{t=1}^T,x^a\}$ and $y=\{y_t\}_{t=1}^T$, respectively.
The inference process of CVAE aims to infer the informative values of the latent variable $z$ given the observed data $X$ and $y$ by computing the posterior $p_\theta(z|X,y)$, which is intractable and usually approximated with the variational posterior $q_\phi(z|X,y)$. $\theta$ and $\phi$ are the parameters of the true posterior and the approximated variational posterior, respectively.
CVAE is trained to find the optimal generation parameters $\theta^*$ and inference parameters $\phi^*$ following the maximum log-likelihood learning pipeline\footnote{Please see the supplementary material for the complete derivation.}:
\begin{equation}
    \{\theta^*,\phi^*\}=\arg\max_{\theta,\phi} \log p_\theta(y|X),
\end{equation}
where the log-likelihood term is achieved as:
\begin{equation}
\begin{aligned}
\label{conditional_likelihood}
    &\log p_\theta(y|X) \\
    &= \underbrace{\mathbb{E}_{q_\phi(z|X,y)}\log p_\theta(y|X,z) - D_{KL}(q_\phi(z|X,y)\|p_\theta(z|X))}_{\text{ELBO}(X,y,\theta,\phi)}\\
    &+D_{\text{KL}}(q_\phi(z|X,y)\|p_\theta(z|X,y)).
\end{aligned}
\end{equation}

By Jensen's inequality, the Kullback–Leibler (KL) divergence term ($D_{KL}$) in Eq.~\eqref{conditional_likelihood} is always greater or equal to zero, thus
maximizing $\log p_\theta(y|X)$
can be achieved by maximizing the evidence lower bound $\text{ELBO}(X,y,\theta,\phi)$:
\begin{equation}
\begin{aligned}
\label{obj_cvae}
        \{\theta^*,\phi^*\} &= \arg\max_{\theta,\phi}\log p_\theta(y|X)\\
        &=\arg\max_{\theta,\phi}\text{ELBO}(X,y,\theta,\phi).
\end{aligned}
\end{equation}

With the reparameterization trick~\cite{VAE1}, the KL term in $\text{ELBO}(X,y,\theta,\phi)$ can be solved in closed-form if both the prior $p_\theta(z|X)$ and posterior $q_\phi(z|X,y)$ are Gaussian.

\noindent\textbf{Multimodal Conditional Variational Auto-encoder.} For the unimodal setting, both $p_\theta(z|X)$ and $q_\phi(z|X,y)$ can be obtained via the reparameterization trick~\cite{VAE1}, leading to closed-form solution of the KL divergence as both $p_\theta(z|X)$ and $q_\phi(z|X,y)$ are Gaussian.
For the multimodal AVS data, the joint posterior 
and joint prior 
need to be estimated before we perform the joint generation process. The conventional solution to model the joint distribution is through the product of experts (PoE)~\cite{product_of_expert,Generalized_Product_of_Experts} or the mixture of experts (MoE)~\cite{shi2019variational}. For the former, the joint distribution is defined as the product of each individual expert, which is Gaussian when each expert is Gaussian, leading to closed form KL computation. However, for PoE, less accurate modeling of one expert will completely destroy the joint distribution modeling. Further, PoE shows limitations in modeling the unimodal contribution due to its multiplicative nature. The additive nature of the MoE makes it effective for the optimization of each individual expert. However, as no closed form exists for the KL term, importance sampling (IS) is usually needed, which is computationally less efficient.

\noindent\textbf{JS Divergence Instead of KL Divergence.} Although MoE is computationally less efficient compared with PoE, its individual modal contribution modeling is attractive. Based on MoE, the $D_{KL}$ term within $\text{ELBO}(X,y,\theta,\phi)$ of Eq.~\eqref{conditional_likelihood} is the lower bound of the weighted sum of individual KLs:
\begin{equation}
\begin{aligned}
\label{kl_lower_bound}
    &D_{KL}(q_\phi(z|X,y)\|p_\theta(z|X))\\
    &\leq \sum_{k=1}^K \phi_k D_{KL}(q_{\phi_k}(z|x^k,y)\|p_\theta(z|x^k)).
\end{aligned}
\end{equation}
where $K$ is the number of modalities, and $\sum_k \phi_k=1$. Although Eq.~\eqref{kl_lower_bound} is effective in providing lower bound with individual modal's distribution for ELBO in Eq.~\eqref{conditional_likelihood}, no joint distribution is involved. Following~\cite{Multimodal_Generative_Learning_Jensen_Shannon_Divergence}, a dynamic prior $f_K$ is introduced, which is the mixture of the involved arguments (individual priors and posteriors), \ie~$f_K$ can be defined as the arithmetic means as in MoE.

With the non-negative nature of KL divergence and the definition of JS divergence, we obtain a new lower bound of ELBO in Eq.~\eqref{conditional_likelihood} as:
\begin{equation}
\small
\begin{aligned}
\label{js_divergence_based}
    &\widehat{\text{ELBO}}(X,y,\theta,\phi)\geq \mathbb{E}_{q_\phi(z|X,y)}\log p_\theta(y|X,z)\\
    &-\sum_{k_1=1}^K\pi_{k_1} D_{KL}(q_\phi(z|X,y)\|f_K)-\sum_{k_2=1}^K\pi_{k_2} D_{KL}(p_\theta(z|X)\|f_K)\\
    &=\underbrace{\mathbb{E}_{q_\phi(z|X,y)}\log p_\theta(y|X,z)-\text{JSD}(q_{\phi}(z|X,y),p_\theta(z|X))}_{\widehat{\text{ELBO}}(X,y,\theta,\phi)},
\end{aligned}
\end{equation}
where $\sum_{k=1}^{2K}\pi_k=1$, and $\text{JSD}$ represents JS divergence. Eq.~\eqref{js_divergence_based} provides
lower bound of ELBO in Eq.~\eqref{conditional_likelihood}, namely $\widehat{\text{ELBO}}(X,y,\theta,\phi)$, which is proven more stable for training~\cite{Multimodal_Generative_Learning_Jensen_Shannon_Divergence}, and robust to noise.

\subsection{Latent Space Factorization}
\label{sec:latent_space_factorization}
Besides stable training and noise robustness, we are also interested in modeling both shared and specific information of the audio-visual input (see Fig.~\ref{fig:model_overview}) to fully explore their contribution.
For each pair of example $(x^k,y)$, where $x^k\!\in\!\{\{x^v_t\}_{t=1}^T,x^a\}$ indexes the modalities with $K=2$ in this paper,
we factorize the latent space $z$ into a modality-shared latent code $c$ and modality-specific latent codes $s^a, s^v$. Then
$\widehat{\text{ELBO}}(X,y,\theta,\phi)$ is re-defined as:\footnote{Please refer to the supplementary material for detailed derivation.}:
\begin{equation}
\textstyle
\small
    \label{final_jensen_shannon_divergence_objective}
    \begin{aligned}
    &\widehat{\text{ELBO}}(X,y,\theta,\phi)\\
    &=\sum_{k=1}^K\mathbb{E}_{q_{\phi_c}(c|X,y)}\left[\mathbb{E}_{q_{\phi_{s^k}}(s^k|x^k,y)}\left[\log p_\theta(y|x^k,s^k,c)\right]\right] \\
    &-\beta \sum_{k=1}^K D_{KL}(q_{\phi_{s^k}}(s^k|x^k,y)||p_\theta(s^k|x^k))\\
    &-\beta \text{JSD}(q_{\phi_{c}}(c|X,y),p_\theta(c|X)),
    \end{aligned}
\end{equation}
where
$q_{\phi_c}(c|X,y)$ and $p_\theta(c|X)$ are the posterior and prior distributions of the shared representation. 
$q_{\phi_{s^k}}(s^k|x^k,y)$ and $p_{\theta_{s^k}}(s^k|x^k)$ are the posterior and prior distribution of the modality-specific latent codes. 
$p_\theta(y|x^k,s^k,c)$ is the prediction generation model. 
All these models can be parameterized by deep neural networks and optimized via stochastic gradient descent. The hyper-parameter $\beta=0.1$ is introduced to achieve stable learning~\cite{higgins2017betavae}.

\noindent\textbf{Efficient Sampling.} The Eq.~\eqref{final_jensen_shannon_divergence_objective} shows that the generation process involves sampling from the joint shared posterior $q_{\phi_c}(c|X,y)$ and posterior of modality-specific latent code $q_{\phi_{s^k}}(s^k|x^k,y)$ of each modality, which is time-consuming. 
In practice, we first perform shared-specific representation fusion, and then we sample latent code from each fused space, achieving efficient sampling. 
Specifically, given the posterior of the latent codes $c\!\sim\! q_{\phi_c}(c|X,y)\!\in\!\mathbb{R}^{T\times L}$, $s^a\!\sim\! q_{\phi_{s^a}}(s^a|x^a,y)\!\in\!\mathbb{R}^{T\times L}$, $s^v\!\sim\! q_{\phi_{s^v}}(s^v|x^v,y)\!\in\!\mathbb{R}^{T\times L}$ ($L$ is the dimension of the latent space),
we concatenate the shared representation with each specific representation to get the fused representation of each modality.
Then we obtain
$sc^a, sc^v$, representing the fused feature of audio and visual data, respectively.
To achieve the reconstruction of $p_\theta(y|x^k,s^k,c)$ in Eq.~\eqref{final_jensen_shannon_divergence_objective}, instead of performing sampling from each specific latent code and shared latent code, we sample from $sc^a, sc^v$, and rewrite the reconstruction term, \ien~the first term in Eq.~\eqref{final_jensen_shannon_divergence_objective}, as:
\begin{equation}
\label{rec_term}
    \mathcal{L}_\text{rec}=\sum_{{sc\in\{sc^a,sc^v\}}}\left[\mathbb{E}\left[\log p_\theta(y|x,sc)\right]\right],
\end{equation}
where $x$ corresponds to the modality of data, \ien~audio or visual, of the current fused latent code $sc$.

\noindent\textbf{Hybrid Loss.} 
As the VAE samples from the posterior for training and prior to testing.
To achieve consistent training and testing, we define a Gaussian stochastic neural network (GSNN)~\cite{cvae_nips} based objective by sampling from the prior distribution as well to avoid the posterior/prior distribution gap, leading to the hybrid
objective as:
\begin{equation}
    \label{hybrid_rec_obj}
    \widehat{\text{HELBO}}(X,y,\theta,\phi)=\alpha_1\widehat{\text{ELBO}}(X,y,\theta,\phi)+(1-\alpha_1)\mathcal{L}_{\text{GSNN}},
\end{equation}
where $\alpha_1\!=\!0.5$ is used to balance the two objectives, $\mathcal{L}_{\text{GSNN}}$ represents the reconstruction term of
Eq.~\eqref{final_jensen_shannon_divergence_objective}, which is achieved
by sampling from 
the prior distribution.

The latent space factorization is effective in generating modality-shared and specific representations.
However, no constraints are applied to the
representations, making it hard to decide the
reliability of the latent codes.
We tackle this issue by proposing a representation orthogonality constraint in Sec.~\ref{orthogonality_constraint} and a shared information completeness regularization in Sec.~\ref{shared_rep_regularization}.


\subsection{Representation Orthogonality Constraint}
\label{orthogonality_constraint}
We introduce a
representation orthogonality constraint to ensure that the modality-shared and modality-specific representations capture different information within each unimodal data.
Specifically, given the latent codes $c, s^a, s^v$, we introduce the difference loss~\cite{domain_separation_network} as:
\begin{equation}
    \label{final_difference_loss}
    \mathcal{L}_{\text{diff}}=\|c^Ts^a\|_F^2 + \|c^Ts^v\|_F^2 + \|(s^a)^T s^v\|_F^2,
\end{equation}
where $\|\cdot\|_F^2$ is the squared Frobenius norm. With the difference loss, we aim to obtain the exclusive feature in each factorized feature representation.

\subsection{Shared-Information Completeness}
\label{shared_rep_regularization}
As discussed in Sec.~\ref{sec:latent_space_factorization}, we fuse the shared representation with each modality-specific representation and obtain $sc^a, sc^v$, representing the task-related information from audio and visual, respectively.
To explicitly model the effectiveness of the fused latent space, we treat them ($sc^a, sc^v$) as two different views of the same target following representation learning~\cite{bachman2019learning,do2021clustering,Learning_View_Disentangled_Human_Pose_Representation_MI}, and introduce
mutual information maximization as a regularizer to measure the shared information between
$sc^a$ and $sc^v$.
Given two random variables $SC^a,SC^v$, the mutual information is defined as:
\begin{equation}
    \label{mutual_info_defination}
    I(SC^a;SC^v)=\mathbb{E}_{p(sc^a,sc^v)}\left[\log \frac{p(sc^a,sc^v)}{p(sc^a)\cdot p(sc^v)}\right].
\end{equation}

According to Bayesian's law, we
obtain the mutual information variational lower bound~\cite{ba_mmm_nips03}, namely $I_\text{ba}$ via:
\begin{equation}
\textstyle
\small
    \label{mutual_info_iba}
    \begin{aligned}
    I(SC^a;SC^v)
    \geq \mathbb{E}_{p(sc^a,sc^v)}[\log q_\kappa(sc^a|sc^v)] + H(SC^a)
    \triangleq I_\text{ba},
    \end{aligned}
\end{equation}
where $q_\kappa(sc^a|sc^v)$ is the variational approximation of
$p(sc^a|sc^v)$. Following~\cite{cheng2020club}, $q_\kappa(sc^a|sc^v)$ is formulated
as a multivariate Gaussian distribution $q_\kappa(sc^a|sc^v)=\mathcal{N}(sc^a|\mu(sc^v),\sigma^2(sc^v)\mathbf{I})$ to predict mean $\mu(sc^v)$ and variance $\sigma^2(sc^v)$, respectively, where each statistic is modeled with two fully connected layers with Tanh activation function in the middle, and $\kappa$ represent parameters of the four fully connected layers.
$H(SC^a)$ is the differential entropy of $SC^a$.
As the audio encoder is fixed in this paper, we choose to simplify the entropy computation and
treat $H(SC^a)$ as a constant~\cite{chen_infogan_NIPS_2016}.

Based on the variational lower bound $I_\text{ba}$, we then define the shared-information completeness loss function as: $\mathcal{L}_\text{sic}=-I_\text{ba}$.
Further, we introduce the hybrid loss function for the posterior and the prior distribution, leading to: 
\begin{equation}
    \label{shared_information_maximization}
    \mathcal{L}_\text{sic} = -\alpha_2 I_\text{ba}^{\text{po}}-(1-\alpha_2) I_\text{ba}^{\text{pr}},
\end{equation}
where $I_\text{ba}^{\text{po}}$ ($I_\text{ba}^{\text{pr}}$) is the lower bound of the mutual information for the posterior (prior) distribution,
and $\alpha_2=0.5$ is introduced to balance the two objectives.

\subsection{The Model}

Four central modules or constraints are included in our framework (see Fig.~\ref{fig:model_overview}), namely: \textbf{1)} \enquote{modal encoding} to extract the feature of each modality; \textbf{2)} \enquote{latent space encoding} for multimodal latent feature representation learning; \textbf{3)} \enquote{decoder} for the segmentation maps prediction; \textbf{4)} \enquote{objective function} for supervised learning and explicit multimodal representation constraints.

\noindent\textbf{Modal Encoding.}
We perform two branches with two encoders to encode the visual and audio data. For the visual branch, we use the ImageNet pre-trained backbone followed by a one-layer convolution as neck to produce the multi-scale visual features $\{V_m\}_{m=1}^4 \in \mathbb{R}^{T\times h_m \times w_m \times C_m}$, where $(h_m, w_m)=(H,W)/2^{m+1}, C_m\!=\!128$. $H,W$ is the spatial resolution of the input video.
We use PVTv2~\cite{wang_Pvtv2_CVM_2022} or ResNet50~\cite{he_resnet_cvpr_2016} as our visual backbone, which keeps the same as AVSBench~\cite{zhou_AVSBench_ECCV_2022}.
For the audio branch, we follow
\cite{zhou_AVSBench_ECCV_2022}, and use a frozen VGGish~\cite{hershey_VGGish_icassp_2017} model pre-trained on AudioSet~\cite{gemmeke_audioset_icassp_2017} to process the spectrogram of input audio to extract audio features $A\!\in\! \mathbb{R}^{T\times d}$, where $d=128$. And $T=5$ denotes the length of the video.
We also
keep the temporal pixel-wise audio-visual interaction (TPAVI)~\cite{zhou_AVSBench_ECCV_2022} module in our framework, which is a cross-modal attention based fusion module that takes visual features as query and value, audio features as key to achieve multi-scale feature fusion in the feature space and obtain $\{\bar{V}_m\}^4_{m=1}$.

\noindent\textbf{Latent Space Encoding.}
The main idea of our proposed method is achieving modality encoding on a reliable latent space (Sec.~\ref{sec:latent_space_factorization}), as shown in Fig.~\ref{fig:model_overview}. We use $\phi_{s^v}, \phi_{s^a}, \phi_{c}$ parameterized by three simple neural networks to get latent feature embedding $s^v, s^a, c \in \mathbb{R}^{T\times L}$ as prior distributions ($L=16$).
For $\phi_{s^v}$, we use five convolutional layers followed by leakyReLU~\cite{maas_leakyrelu_icml_2013} with two fully connected layers to encode the input video sequence.
While for $\phi_{s^a}$, we employ two fully connected layers to map the VGGish encoded audio features into the latent space. 
Further, video sequence and audio features are fed into $\phi_{c}$ jointly, thus we perform late fusion on the audio features and five convolutional layers followed by leakyReLU encoded visual features to obtain the joint distribution from the fused features.
For the posterior network, we design three networks with the same structure and take the segmentation maps $y$ as input by concatenating the video sequence and segmentation maps. We omit drawing the posterior space in Fig.~\ref{fig:model_overview} for easy viewing.

\noindent\textbf{Decoder.}
We adopt the decoder of Panoptic-FPN~\cite{kirillov_panopticfpn_cvpr_2019} to decode the final segmentation maps for its flexibility and effectiveness, which is the same as AVSBench~\cite{zhou_AVSBench_ECCV_2022}. We expand $sc^a, sc^v$ to feature map of the same spatial size as $\bar{V}_4$ by adding two-dimensional gaussian noise with the tile operation, \enquote{sampling} is used to indicate this process in Fig.\ref{fig:model_overview}.
The decoder takes both the deterministic features $\{\bar{V}_m\}^4_{m=1}$ produced by the TPAVI~\cite{zhou_AVSBench_ECCV_2022} module and the expanded latent codes $sc^a, sc^v$ from the fused latent space as input to produce the final segmentation maps.


\noindent\textbf{Objective Function.}
As discussed above, our final objective function contains the optimization of the evidence lower bound and the practical constraints for latent space representation, and it can be defined as,
\begin{equation}
\begin{aligned}
\label{main_loss}
  \mathcal{L} & = -\widehat{\text{HELBO}}(X,y,\theta,\phi)+\lambda_1\mathcal{L}_\text{diff} \\
              & + \lambda_2\mathcal{L}_\text{sic} + \lambda_3\mathcal{L}_\text{AVM},
\end{aligned}
\end{equation}
where $\widehat{\text{HELBO}}(X,y,\theta,\phi)$ indicates the lower bound for our proposed ECMVAE optimization, which is defined in detail in Eq.~\eqref{hybrid_rec_obj} and Eq.~\eqref{final_jensen_shannon_divergence_objective}. 
We use a weighted structure-aware function~\cite{wei_F3Net_AAAI_2020} to compute the hybrid reconstruction part in $\widehat{\text{HELBO}}(X,y,\theta,\phi)$.
$\mathcal{L}_\text{diff}$ and $\mathcal{L}_\text{sic}$ are the orthogonality constraint and the shared-information completeness loss defined in Eq.~\eqref{final_difference_loss} and Eq.~\eqref{shared_information_maximization}.
$\mathcal{L}_\text{AVM}$ indicates the audio-visual mapping loss proposed by~\cite{zhou_AVSBench_ECCV_2022} as a regularization term to promote the similarity between the audio-visual features.
Empirically, we set the hyper-parameters $\{\lambda_1, \lambda_2, \lambda_3\}$ as $\{0.001,0.01,0.5\}$ for balanced training. 

\section{Experimental Results}
\subsection{Implementation Details}
\noindent\textbf{Datasets.} We conduct experiments on the AVSBench~\cite{zhou_AVSBench_ECCV_2022} dataset, which contains 5,356 video sequences with corresponding audio data and binary per-pixel annotations. 
Each video in the dataset contains five frames, extracted separately from a five-second video, where the audio length is also five seconds. 
This dataset contains two settings, named S4 and MS3, for semi-supervised Single Sound Source Segmentation with only the first frame labeled, and fully supervised Multiple Sound Source Segmentation with all frames labeled. 
The evaluation is done for the entire five frames of the video under both S4 and MS3 settings on the test set.

\begin{figure*}[!htp]
\begin{center}
\includegraphics[width=0.98\linewidth]{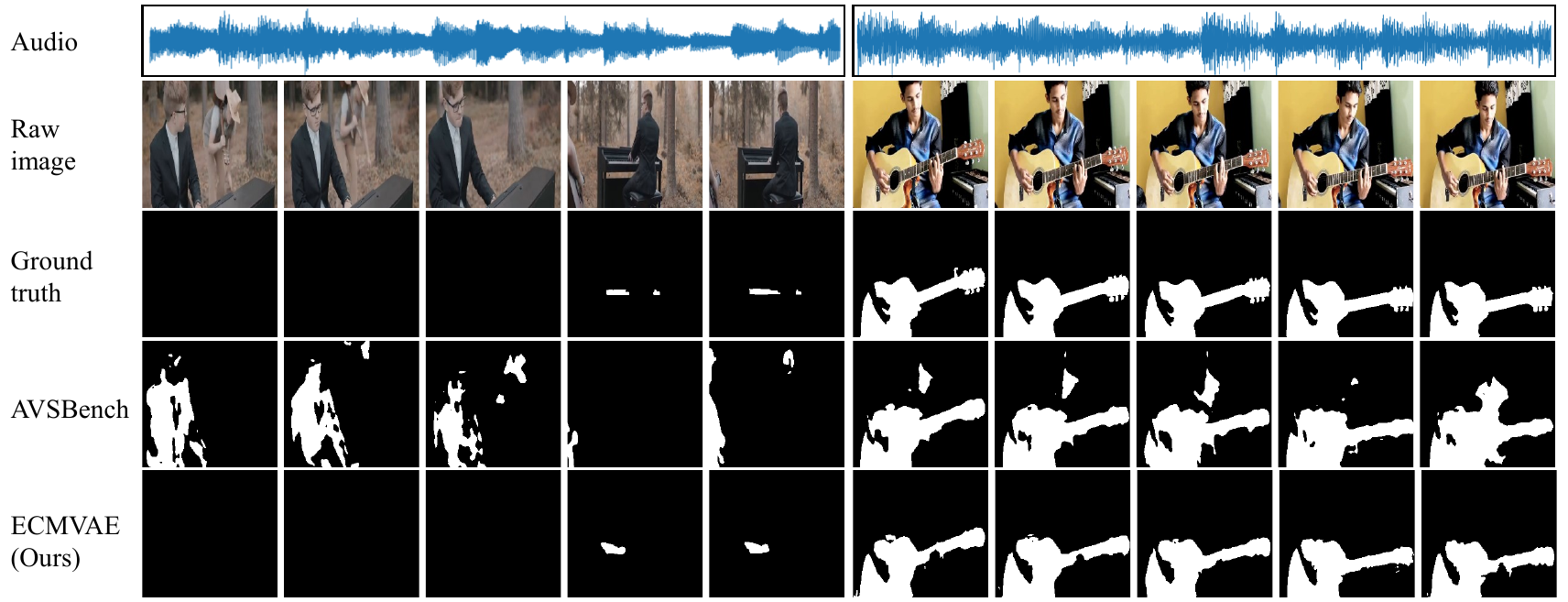}
\end{center}
\vspace{-0.0mm}
\caption{\textbf{Qualitative comparison} between our proposed ECMVAE and AVSBench~\cite{zhou_AVSBench_ECCV_2022}. Our method competently achieves high segmentation performance with better audio temporal and spatial localization quality and detail handling.}
\vspace{-2.0mm}
\label{fig:main_compare}
\end{figure*}

\begin{table}[!htp]
    \caption{\textbf{Quantitative results on the AVSBench dataset~\cite{zhou_AVSBench_ECCV_2022}} in terms of mIOU and F-score under S4 and MS3 settings. We both report the performance with R50 and PVT as a backbone for the results of AVSBench~\cite{zhou_AVSBench_ECCV_2022} and Ours.}
    \label{tab:main_results_on_avsbench}
    \centering
    \small
    \vspace{2.0mm}
    \setlength{\tabcolsep}{1.2mm}{
        \begin{threeparttable}
        \begin{tabular}{cccccc}
        \toprule
        & \multirow{2}{*}{Methods} & \multicolumn{2}{c}{S4}                      & \multicolumn{2}{c}{MS3}                     \\
        \cmidrule(r){3-4}  \cmidrule(r){5-6}
                     &         & mIoU                 & F-score              & mIoU                 & F-score   \\
        \midrule
        \multirow{2}{*}{VOS}& 3DC~\cite{mahadevan_3DC_VOS_2020}  & 57.10   & 0.759   & 36.92   & 0.503    \\
        & SST~\cite{duke_sstvos_cvpr_2021}                       & 66.29   & 0.801   & 42.57   & 0.572    \\
        \midrule
        \multirow{2}{*}{SOD}& iGAN~\cite{mao_transformerSOD_2021}& 61.59   & 0.778   & 42.89   & 0.544    \\
        & LGVT~\cite{zhang_ebm_sod_nips_2021}                    & 74.94   & 0.873   & 40.71   & 0.593    \\
        \midrule
        &AVSBench (R50)~\cite{zhou_AVSBench_ECCV_2022}   & 72.79   & 0.848   & 47.88   & 0.578    \\
        \multirow{2}{*}{AVS}&AVSBench (PVT)~\cite{zhou_AVSBench_ECCV_2022}   & 78.74   & 0.879   & 54.00   & 0.645    \\
         & Ours (R50)               & 76.33          & 0.865          & 48.69          & 0.607           \\
         & Ours (PVT)               & \textbf{81.74} & \textbf{0.901} & \textbf{57.84} & \textbf{0.708}  \\
        \toprule
        \end{tabular}
        \end{threeparttable}
    }
    \vspace{-2.0mm}
\end{table}

\noindent\textbf{Training Details.} 
We conduct experiments on Pytorch~\cite{paszke_pytorch_NIPS_2019} with a single NVIDIA A100 GPU.
The Adam~\cite{Kingma_Adam_ICLR_2015} solver is used to optimize our network with a learning rate of $1\times 10^{-4}$. The training batch size is set to 4. 
We train the network on the S4 subset for 15 epochs, and on the MS3 subset for 30 epochs.
For the MS3 setting, we use all ground-truth of the five frames to build the posterior latent space of our model. While for S4, we repeat the ground-truth of the first frame five times to build the posterior latent space.


\subsection{Comparison with Baseline Methods}
\noindent\textbf{Quantitative Comparison.} Follow the comparison settings with AVSBench~\cite{zhou_AVSBench_ECCV_2022}, we compare the performance of our ECMVAE with baseline AVS models and other related tasks, including video object segmentation (VOS) and salient object detection (SOD). The performance on Mean Intersection over Union (mIoU) and F-score is reported in Table~\ref{tab:main_results_on_avsbench}. 
It can be observed that our method consistently achieves significantly superior segmentation performance than the state-of-the-art methods, especially with 3.00 and 3.84 higher mIOU than the previous AVS method~\cite{zhou_AVSBench_ECCV_2022}, at the S4 and MS3 settings with PVTv2 (\enquote{PVT}) backbone. 
There is also a consistent performance improvement with ResNet (\enquote{R50}) backbone. 
The performance gain comes from our designed multimodal VAE with explicit constraints for representation learning.
Our method also significantly outperforms the VOS and SOD methods, demonstrating the addition of audio information to the segmentation performance. 

\noindent\textbf{Qualitative Comparison.} We provide a qualitative comparison between our proposed method and~\cite{zhou_AVSBench_ECCV_2022} in Fig.~\ref{fig:main_compare}. Our proposed ECMVAE provides a better audio temporal and spatial localization quality, leading to better segmentation performance, especially for the left samples, in localization of the \emph{piano keys}, which is not salient but producing a sound in this scene. 
Our method also achieves better segmentation performance for background noise handling and richer foreground details in the right samples in Fig.~\ref{fig:main_compare}.

\subsection{Ablation Studies}
\label{ablation}
We conduct ablation studies of our proposed method. 
All variations are trained with the PVT backbone.

\begin{table}[!htp]
    \caption{\textbf{Ablation of the VAE based multimodal learning.} We implement a \enquote{CVAE} without audio and \enquote{CMVAE} with audio-visual joint distribution estimation.}
    \vspace{2.0mm}
    \label{tab:ablation_on_mmvae}
    \centering
    \small
    \setlength{\tabcolsep}{1.5mm}{
        \begin{threeparttable}
        \begin{tabular}{ccccc}
        \toprule
        \multirow{2}{*}{Methods} & \multicolumn{2}{c}{S4}    & \multicolumn{2}{c}{MS3}         \\
        \cmidrule(r){2-3}  \cmidrule(r){4-5}
                             & mIoU    & F-score & mIoU    & F-score        \\
        \midrule
        \cite{zhou_AVSBench_ECCV_2022} w/o audio & 77.80   & -       & 48.20   & -        \\
        CVAE                                     & 78.12   & 0.878   & 49.26   & 0.643    \\
        CMVAE                                    & \textbf{80.05} & \textbf{0.889} & \textbf{54.99} & \textbf{0.653}    \\
        \toprule
        \end{tabular}
        \end{threeparttable}
    }
    \vspace{-2.0mm}
\end{table}

\noindent\textbf{Multimodal VAE.}
We explore the effectiveness of the Multimodal VAE in
Table~\ref{tab:ablation_on_mmvae}. Firstly, 
we remove the audio part of the model and disable the TPAVI module to explore the importance of the audio information, leading to
a simple unimodal CVAE~\cite{zhang_ucnet_CVPR_2020} framework with only video input, which is denoted as \enquote{CVAE}. 
For comparison, we implement a \enquote{CMVAE} using the audio signal but without the latent space factorization and our proposed constraints. The better performance of \enquote{CMVAE} compared with \enquote{CVAE} indicates the importance of audio for
AVS,
especially when multiple sound sources exist. We also brought the ablation result of the model without the audio input from~\cite{zhou_AVSBench_ECCV_2022} and compare it with \enquote{CVAE}. The results show that in the absence of audio signals, the VAE structure can still improve segmentation performance, due to the ability of VAE to model the latent space of visual features.


\begin{table}[!htp]
    \caption{\textbf{Ablation of the latent space factorization.} \enquote{Model} indicates using VAE for latent factorization or using AE for feature factorization. \enquote{Factor.} denotes whether using factorization. \enquote{dim.} represents the size of the latent dimensions.}
    \vspace{2.0mm}
    \label{tab:ablation_on_factor}
    \centering
    \small
    \setlength{\tabcolsep}{1.5mm}{
        \begin{threeparttable}
        \begin{tabular}{ccccccc}
        \toprule
        &  &  & \multicolumn{2}{c}{S4}    & \multicolumn{2}{c}{MS3}         \\
        \cmidrule(r){4-5}  \cmidrule(r){6-7}
        Model   & Factor.       & dim.        & mIoU    & F-score & mIoU    & F-score        \\
        \midrule
        \multirow{3}*{VAE}    & -          & 16  & 80.05   & 0.889   & 54.99   & 0.653    \\
                              & -          & 48  & 80.13   & 0.890   & 55.09   & 0.657    \\
                              & \checkmark & 16  & \textbf{80.78} & \textbf{0.893} & \textbf{56.38} & \textbf{0.676} \\
        \midrule
        \multirow{2}*{AE}& -          &-    & 78.74   & 0.879   & 54.00   & 0.645    \\
                              & \checkmark &-    & 78.92   & 0.881   & 54.82   & 0.651    \\
        \toprule
        \end{tabular}
        \end{threeparttable}
    }
    \vspace{-2.0mm}
\end{table}

\begin{figure*}[!htp]
\begin{center}
\includegraphics[width=0.9\linewidth]{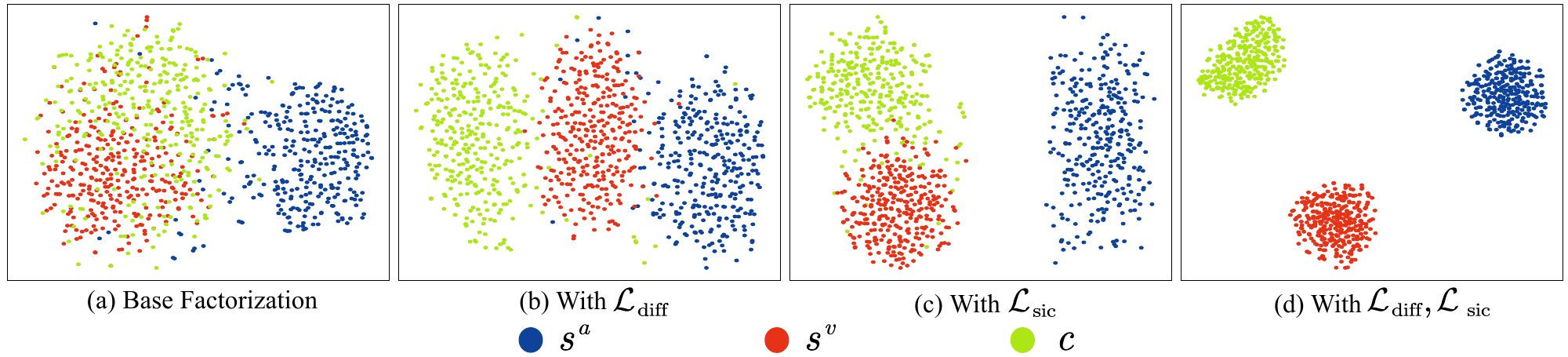}
\end{center}
\vspace{-2.0mm}
\caption{\textbf{Visualization of the modality-shared and modality-specific latent codes} ($s^v, s^a, c$) in the MS3 testing set using t-SNE~\cite{van_tsne_jmlr_2008} projections. Best viewed on screen.}
\label{fig:shared_specific_latent_feat}
\vspace{-2.0mm}
\end{figure*}

\noindent\textbf{Latent Space Factorization.}
As described in Sec.~\ref{sec:latent_space_factorization}, we factorize the latent space of multimodal VAE into modality-shared ($c$) and modality-specific ($s^v$ and $s^a$) representations. 
As reported in Table~\ref{tab:ablation_on_factor}, removing the latent space factorization leads to obvious performance degradation.
We also train a non-factorized model with $2\times$ larger latent dimensions, which holds comparable latent space capacities with the factorized model. 
It can be seen that the performance gain from the larger latent space dimensions is not as obvious as the factorization strategy.
Further, we compare feature factorization (\enquote{AE}) on the feature space of~\cite{zhou_AVSBench_ECCV_2022} with our proposed latent space factorization (\enquote{VAE}). Table~\ref{tab:ablation_on_factor} shows that only performing factorization on a semantic meaningful and continuous latent space, \ie~via using VAE~\cite{VAE1}, can achieve larger performance improvements.

\begin{table}[!htp]
    \caption{\textbf{Ablation of the JS Divergence.} We implement \enquote{PoE} and \enquote{MoE} with KL divergence, and a simple \enquote{KL} model.}
    \vspace{2.0mm}
    \label{tab:ablation_on_js}
    \centering
    \small
    \setlength{\tabcolsep}{1.5mm}{
        \begin{threeparttable}
        \begin{tabular}{ccccc}
        \toprule
        \multirow{2}{*}{Methods} & \multicolumn{2}{c}{S4}    & \multicolumn{2}{c}{MS3}         \\
        \cmidrule(r){2-3}  \cmidrule(r){4-5}
                & mIoU    & F-score & mIoU    & F-score        \\
        \midrule
        KL      & 80.78   & 0.893   & 56.38   & 0.676    \\
        PoE     & 81.38   & 0.894   & 57.35   & 0.688    \\
        MoE     & 81.49   & 0.897   & 57.53   & 0.694    \\
        JS      & \textbf{81.74}   & \textbf{0.901}   & \textbf{57.84}   & \textbf{0.708}    \\
        \toprule
        \end{tabular}
        \end{threeparttable}
    }
    \vspace{-2.0mm}
\end{table}

\noindent\textbf{JS Divergence.}
We conduct experiments of PoE and MoE with KL divergence to show the trade-off of JS divergence between the computational efficiency of the inference process and the predictive quality of the generation process. The performance in Table~\ref{tab:ablation_on_js} shows the effectiveness of JS divergence on both S4 and MS3 settings. Note that, \enquote{PoE}, \enquote{MoE} and \enquote{JS} are based on our formulation of the task, which has not been explored in the AVS area.

\begin{table}[!htp]
    \caption{\textbf{Ablation of the latent space constraints.}
    $\mathcal{L}_\text{diff},\mathcal{L}_\text{sic}$ indicate our proposed loss functions for multimodal learning.}
    \vspace{2.0mm}
    \label{tab:ablation_on_loss}
    \centering
    \small
    \setlength{\tabcolsep}{1.8mm}{
        \begin{threeparttable}
        \begin{tabular}{cccccc}
        \toprule
        & & \multicolumn{2}{c}{S4}    & \multicolumn{2}{c}{MS3}         \\
        \cmidrule(r){3-4}  \cmidrule(r){5-6}
        $\mathcal{L}_\text{diff}$ & $\mathcal{L}_\text{sic}$   & mIoU    & F-score & mIoU    & F-score        \\
        \midrule
        -          & -          & 81.09  & 0.895  & 57.01   & 0.684 \\
        \checkmark & -          & 81.51  & 0.899  & 57.65   & 0.692 \\
        -          & \checkmark & 81.47  & 0.898  & 57.51   & 0.694 \\
        \checkmark & \checkmark & \textbf{81.74}  & \textbf{0.901}  & \textbf{57.84}   & \textbf{0.708} \\
        \toprule
        \end{tabular}
        \end{threeparttable}
    }
    \vspace{-5.0mm}
\end{table}


\noindent\textbf{Orthogonality Constraint.}
As reported in Table~\ref{tab:ablation_on_loss}, the orthogonality constraint provides 0.64 mIOU gain under the MS3 setting, which facilitates the latent space factorization.
We also perform t-SNE~\cite{van_tsne_jmlr_2008} projection to visualize the factorized latent code with and without such constraint. 
As compared between Fig.~\ref{fig:shared_specific_latent_feat} (a) and (b), the three latent codes in the latent space are divided into different subspaces to ensure that each latent code encodes different information.

\noindent\textbf{Mutual Information Maximization.}
As compared in Table~\ref{tab:ablation_on_loss}, the mutual information maximization by $\mathcal{L}_\text{sic}$ also serves a crucial impact for the explicit constraint of the shared latent space and improves the mIOU from 57.51 to 57.84.
Since the $\mathcal{L}_\text{sic}$ maximizes the mutual information between $sc^v$ and $sc^a$, which are fused from the factorization latent codes $s^v, s^a, c$. This increases the amount of information contained in latent space and makes it more effective for facilitating factorization. Fig.~\ref{fig:shared_specific_latent_feat} also confirms this view and demonstrates the effectiveness of $\mathcal{L}_\text{sic}$ in achieving effective \enquote{multi-view} representation learning~\cite{bachman2019learning,do2021clustering,Learning_View_Disentangled_Human_Pose_Representation_MI}.


\subsection{Analysis}
\noindent\textbf{Pre-training on the Single-source Subset.}
In AVSBench~\cite{zhou_AVSBench_ECCV_2022}, they conducted experiments by initializing model parameters by pre-training on S4 dataset.
We also perform
experiments with such setting (see Table~\ref{tab:results_for_pertrain}).
We can observe that both our method and AVSBench~\cite{zhou_AVSBench_ECCV_2022} can benefit from the model pre-trained on S4. The PVT-based model can gain 2.97\% mIOU performance by such a strategy and reach 60.81\% mIOU. The pre-training strategy can bring more significant improvements (8.87\% mIOU) to ResNet50-based models and reach 57.56\% mIOU, which is even higher than PVT-based models (57.34\%).

\begin{table}[!htp]
    \caption{\textbf{Performance comparison with different initialization strategies for MS3 dataset.}
    As AVSBench~\cite{zhou_AVSBench_ECCV_2022} does not report its F-score in the paper, we only report its mIOU.
    The values in parentheses indicate the performance improvement based on S4 pre-training compared with training from scratch.
    }
    \vspace{2.0mm}
    \label{tab:results_for_pertrain}
    \centering
    \small
    \setlength{\tabcolsep}{1.4mm}{
        \begin{threeparttable}
        \begin{tabular}{ccccc}
        \toprule
        \multirow{2}{*}{Methods} & \multicolumn{2}{c}{From scrach}    & \multicolumn{2}{c}{Pre-trained on S4}         \\
        \cmidrule(r){2-3}  \cmidrule(r){4-5}
                                                        & mIoU    & F-score & mIoU    & F-score        \\
        \midrule
        AVSBench (R50)~\cite{zhou_AVSBench_ECCV_2022}   & 47.88   & - & 54.33 {\scriptsize ($\uparrow$ 6.45)}   & -  \\
        AVSBench (PVT)~\cite{zhou_AVSBench_ECCV_2022}   & 54.00   & - & 57.34 {\scriptsize ($\uparrow$ 3.34)}   & -  \\
        Ours (R50)  & 48.69          & 0.607          & 57.56 {\scriptsize ($\uparrow$ 8.87)}   & 0.674           \\
        Ours (PVT)  & \textbf{57.84} & \textbf{0.708} & \textbf{60.81} {\scriptsize ($\uparrow$ 2.97)} & \textbf{0.729}   \\
        \toprule
        \end{tabular}
        \end{threeparttable}
    }
    \vspace{-2.0mm}
\end{table}

\begin{table}[!htp]
    \caption{\textbf{Parameters and inference time.}
    }
    \vspace{2.0mm}
    \label{tab:param_and_time}
    \centering
    \small
    \setlength{\tabcolsep}{0.95mm}{
        \begin{threeparttable}
        \begin{tabular}{ccccc}
        \toprule
        \multirow{2}{*}{Methods} & \multicolumn{2}{c}{R50}    & \multicolumn{2}{c}{PVT}         \\
        \cmidrule(r){2-3}  \cmidrule(r){4-5}
                                                  & Param. (M)    & Time (ms)  & Param. (M)    & Time (ms)        \\
        \midrule
        AVSBench~\cite{zhou_AVSBench_ECCV_2022}   & 70.50   & 28  & 101.32   & 53    \\
        Ours         & \textbf{33.97} & \textbf{23} & \textbf{91.18}     & \textbf{46}   \\
        \toprule
        \end{tabular}
        \end{threeparttable}
    }
    \vspace{-2.0mm}
\end{table}

\noindent\textbf{Parameters and Efficiency.}
In Table~\ref{tab:param_and_time}, we compare the parameters and inference time between ours and AVSBench. Note that although posterior nets and prior nets are used in our framework, as
all the latent space encoders are quite lightweight (1M), thus our model capacity will not change significantly.
Moreover, we replace the neck from ASPP~\cite{chen_deeplabv3_2017rethinking} to one-layer convolution and reduce the number of neck channels from 256 to 128, which leads to smaller parameter numbers and faster inference speed.

\noindent\textbf{Limitations.} Similar to the other VAE~\cite{VAE1} based solutions, our model also suffers from the risk of posterior collapse, where the posterior of the latent variable is equal to prior~\cite{bowman2015generating,van2017neural}, making $y$ in our case not encoded in the latent variables. To avoid such phenomenon, contrastive learning~\cite{chopra2005learning,dimension_reduction_lecun} can be studied to learn more compact features of each modality or score based diffusion models~\cite{diffusion_model_raw,denoising_diffuion,song2019generative,score_based_model_latent_space} can be investigated for more informative latent space. 


\section{Conclusion}
We have worked on
audio-visual segmentation (AVS), aiming to segment the sound producers of the scene. As audio data can be treated as \enquote{command}, we argue extensive exploration of audio is critical for effective AVS.
Inspired by this observation,
we have introduced an Explicit Conditional Multimodal Variational Auto-Encoder (ECMVAE) for an audio-visual segmentation model with latent space factorization with explicit constraints to extensively explore the shared and specific representations of audio and visual data.
Extensive experimental results verify the effectiveness of our proposed framework.
Although we focused on AVS with two modalities, the proposed framework can be extended to more modalities and other audio-visual collaboration tasks~\cite{tian_avpcite1_eccv_2020, yu_avel_cross_atten1_mm2022, arandjelovic_look_listen_and_learn_CVPR_2017}.

\section{Acknowledgments}
This research was supported in part by the National Natural Science Foundation of China (62271410), and the Fundamental Research Funds for the Central Universities.


{\small
\bibliographystyle{unsrt}
\bibliography{egbib}
}

\end{document}